\documentclass[journal]{IEEEtran}
\usepackage{amsmath}
\usepackage{amsthm}
\usepackage{amssymb}
\usepackage{cite}
\usepackage{graphicx}
\usepackage{graphics}
\usepackage{epstopdf}
\usepackage{color}
\usepackage{soul}
\usepackage{xcolor}
\usepackage{bm}
\usepackage{hhline}
\usepackage{balance}
\usepackage{multirow}
\usepackage{booktabs}
\usepackage{subfigure}
\usepackage[colorlinks,linkcolor=blue]{hyperref}
\usepackage[ruled]{algorithm2e}
\usepackage{silence}
\usepackage{bbding}
\usepackage{tasks}
\usepackage{footnote}
\usepackage{fancyhdr}
\settasks{
  style=itemize,
  after-item-skip=0pt
}
\newtheoremstyle{mythm}{1.5ex plus 1ex minus .2ex}{1.5ex plus 1ex minus .2ex}{}{}{\itshape}{:}{5pt plus 1pt minus 1pt}{}
\theoremstyle{mythm}

\allowdisplaybreaks

\graphicspath{{image/}}

\begin{document}
\title{TND-NAS: Towards Non-Differentiable Objectives in Differentiable Neural Architecture Search}
\author{Bo Lyu, Shiping Wen
\thanks{
}
\thanks{B. Lyu is with School of Computer Science and Engineering, University of Electronic Science and Technology of China (email: blyucs@outlook.com). S. Wen and Z. Yan are with Australian AI Institute, Faculty of Engineering and Information Technology, University of Technology Sydney, Ultimo 2007, Australia (email: \{shiping.wen; yan.zheng\}@uts.edu.au).}
}
{}
\maketitle
\begin{abstract}
Differentiable architecture search has gradually become the mainstream research topic in the field of Neural Architecture Search (NAS) for its high efficiency compared with the early NAS (EA-based, RL-based) methods.
Recent differentiable NAS also aims at further improving the search performance and reducing the GPU-memory consumption. However, these methods are no longer naturally capable of tackling the non-differentiable objectives, e.g., energy, resource-constrained efficiency, and other metrics, let alone the multi-objective search demands. Researches in the multi-objective NAS field target this but requires vast computational resources cause of the sole optimization of each candidate architecture. In light of this discrepancy, we propose the \textit{TND-NAS}, which is with the merits of the high efficiency in differentiable NAS framework and the compatibility among non-differentiable metrics in Multi-objective NAS. Under the differentiable NAS framework, with the continuous relaxation of the search space, \textit{TND-NAS} has the architecture parameters ($\alpha$) been optimized in discrete space, while resorting to the progressive search space shrinking by $\alpha$. Our representative experiment takes two objectives (\textit{Parameters}, \textit{Accuracy}) as an example, we achieve a series of high-performance compact architectures on CIFAR10 (1.09M/3.3\%, 2.4M/2.95\%, 9.57M/2.54\%) and CIFAR100 (2.46M/18.3\%, 5.46/16.73\%, 12.88/15.20\%) datasets. Favorably, compared with other multi-objective NAS methods, \textit{TND-NAS} is less time-consuming (1.3 GPU-days on \textit{NVIDIA 1080Ti}, 1/6 of that in NSGA-Net).
\end{abstract}

\begin{IEEEkeywords}
Neural architecture search, reinforcement learning, non-differentiable, supernetwork
\end{IEEEkeywords}

\section{Introduction}
\IEEEPARstart{N}{eural} Architecture Search aims at alleviating the tremendous labor of manual tuning on neural network architectures, which has facilitated the development of AutoML \cite{guo2020hierarchical,stamoulis2020single,he2021automl}. 
Recently, under the fast-growing in this area, NAS models have surpassed previous manually designed models in various research fields.
A mass of computational overhead (electricity cost, time cost) is required (20,000 GPU-days in \cite{zoph2017neural} and 2,000 in \cite{zoph2018learning}), which has beyond the reach of ordinary research institutes and commercial organizations. So some subsequent literature try to promote the efficiency of the search procedure, e.g. ENAS \cite{pham2018efficient}.
However, due to the efficiency issue, less research attention is addressed on the RL-based NAS approaches after the rise of differential NAS \cite{liu2018darts}.

By continuous relaxation of the search space, differentiable NAS researches make the loss function differentiable w.r.t architecture weights, thus the search can be processed directly by gradient-based optimization. Benefits from the weight-sharing of different candidate architectures that come from the unified supernetwork \cite{Brock2017SMASH,bender2018understanding,lyu2021neural}, it saves the unnecessary time-cost and computation-consumption that are resulted from the candidate model's sole training from scratch.
Even with high search efficiency, differentiable NAS researches rarely involves the non-differentiable objectives, e.g., energy, latency, or memory consumption, and multi-objective are merely jointly considered.

Parallel to the explosive development of the differentiable NAS sub-field, the multi-objective NAS methods (MnasNet \cite{tan2019mnasnet}, DPP-Net \cite{dong2018dpp-net}, MONAS \cite{hsu2018monas}, Pareto-NASH \cite{elsken2018moas}, \cite{lyu2021multiobjective}, \cite{lu2019nsga}) dedicate to searching for the neural architectures in discrete space with the consideration of multi-dimensional metrics.
There is no doubt that the computational overhead of the multi-objective NAS is huge.

\begin{figure*}[!htb]
    \centering
        \includegraphics[width=17cm]{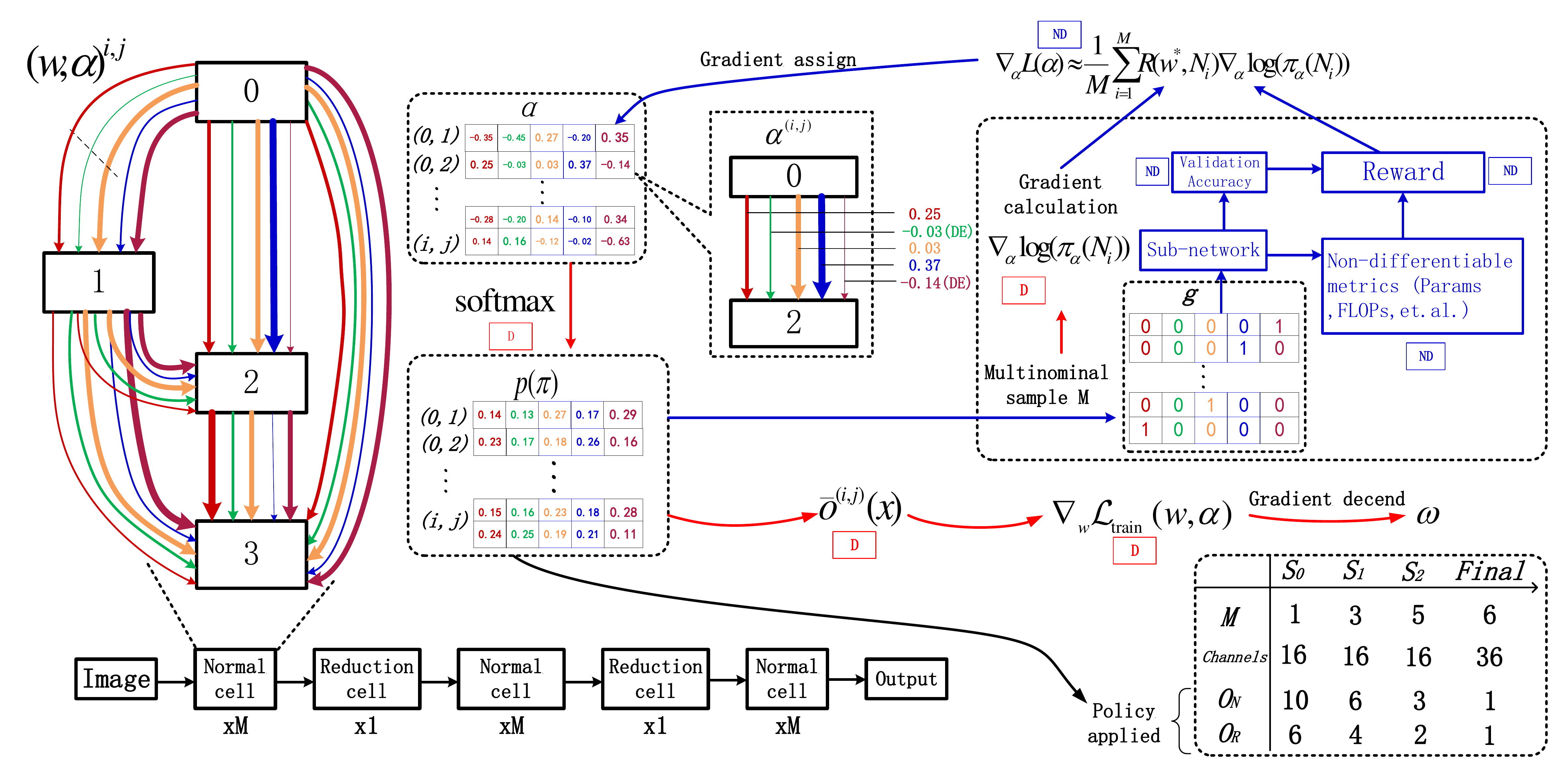}
    \caption{Illustration of the overall framework of TND-NAS. 
    }
    \label{fig:Illustration}
\end{figure*}

Our method relies on the differentiable NAS as the main search framework. In the meantime, the depth of the model is increased gradually \cite{chen2019progressive}. The difference is that we detach the architecture parameter $\alpha$ from the gradient descend framework and formulate the training as an optimization problem in discrete space by the reinforcement learning algorithm. In this way, the non-differentiable metrics can naturally be involved in the search process.
Generally, our work jointly searches the architectures across the differentiable and non-differentiable metrics, it combines the merits of both differentiable NAS and multi-objective NAS. We name our search framework as ``TND-NAS'', the overall structure is illustrated in Fig. \ref{fig:Illustration}. Though under differentiable NAS framework, TND-NAS has the architecture parameters ($\alpha$) been optimized in discrete space, and resort to the progressive search strategy by $\alpha$ to shrink the search space.

The contributions of this work may be summarized as follows:\\
1. Our proposed TND-NAS methods is capable to tackle the non-differentiable objectives under the differentiable search framework, that is, with the merits of high efficiency in differentiable NAS and the objective compatibility of multi-objective NAS.\\
2. We comprehensively address the ``optimization gap'', ``depth gap'' issues, as well as the ``GPU-memory consumption'' issue in our search. Our search process is also end-to-end, that is, without any skip-connection dropout and compulsive restriction of node indegree.\\
3. Through flexible and customized search configurations, the visualization of architecture evolving during the search process show that our method can reach the trade-off among differentiable and non-differentiable metrics.\\

\section{Related Work}
\textbf{RL/EA-based NAS.}
Traditional architecture search methods start from employing Evolutionary Algorithm (EA) as the search strategy \cite{angeline1994evolutionary,stanley2002evolving, sun2019evolving, sun2020automatically, real2019regularized, liu2017hierarchical, lu2019nsga}. In these works, high-performing network architectures are mutated, and less promising architectures are discarded.
Recently, the significant success of RL-based NAS is first reported by \cite{zoph2017neural, zoph2018learning}. These works provide remarkable results on CIFAR-10 and PTB datasets, but require excessive computational resources. ENAS \cite{pham2018efficient} comes in the continuity of previous work \cite{zoph2017neural,zoph2018learning}, it proposes the weight-sharing strategy to significantly improve the searching efficiency. 

\textbf{Differentiable NAS.}
P-DARTS \cite{chen2019progressive} introduces the progressive NAS which evolves the supernetwork from a shallow-wide one to a deep-narrow one. ProxylessNAS \cite{cai2018proxylessnas} also employs continuous relaxation in DARTS and try to directly search without proxy-task (targeting the ``proxy-task gap'').
And benefit from the regularity of pre-defined chain-style structures, ProxylessNAS constructs \textit{inference latency} expectation of each chain-style layer by accumulating each layer's predicted latency value, and formulating it as the regularization item. This makes \textit{inference latency} metric differentiable, thus the bi-objective problem may be optimized uniformly. Whereas the pre-defined chain-style network limits its search space and results in expensive GPU memory and time cost.
Further, for the non-chain-style backbone, method in ProxylessNAS is not feasible to tackling the \textit{Latency}, \textit{FLOPs}, and \textit{Parameters}, as these metrics cannot be calculated by the linear transformation function.

\textbf{Multi-objective NAS and Platform-aware NAS.}
Multi-objective NAS methods (MnasNet \cite{tan2019mnasnet}, MONAS \cite{hsu2018monas}) concentrate on searching the architectures with consideration of multi-dimensional evaluation metric, and 
In these researches, model's consumption (Parameters, FLOPs, Latency, Energy) are jointly employed to formulate the reward-penalty coefficients of the \textit{Accuracy} \cite{cheng2018searching}. Through the optimization of the single-policy multi-objective reinforcement learning algorithm, the search process is prone to emit the candidate architectures that reach the trade-off among these contradictory metrics. 
DPP-Net \cite{dong2018dpp-net} proposed the progressive search methods that involve device-aware characteristics, including QoS (Quality of Service) and hardware resource requirements (e.g., memory size), which are critical metrics of the deployment of deep neural network. It also takes into consideration the different target platforms, e.g. workstations, mobiles, and embedded devices. In LEMONADE \cite{elsken2019efficient}, an evolutionary algorithm is proposed to address the multi-objective NAS. It achieves state-of-the-art Pareto-optimal performance on CIFAR10. Critically, these multi-objective NAS methods are based on reinforcement learning or evolutionary algorithms, so their disadvantages in search efficiency are obvious, e.g., LEMONADE \cite{elsken2019efficient} consumes 80 GPU-days.
Also, dedicating to the customized NAS, some platform-aware methods \cite{lyu2021resource} are proposed, as well as the method that is specified for the resource-constrained situation \cite{lu2021reducing}. 

\textbf{One-shot NAS.}
One-shot NAS researches \cite{Brock2017SMASH, bender2018understanding,guo2019single} construct the search space into a unified supernetwork, from which sub-architectures are solely sampled and evaluated, and these sub-architectures share the common weights of the supernetwork (directly inherited).
Before the sampling in one-shot NAS, the supernetwork is trained uniformly without any bias to any sub-networks, but intuitively, a well-trained supernetwork does not always produce the best-performance sub-network. This brings up the ``optimization gap'' between the sub-networks and supernetwork. As is discussed in \cite{xie2021weight}, ``optimization gap'' means that a well-optimized super-network does not necessarily produce well-performance sub-architectures, whereas this is exactly what the conventional supernetwork-based NAS method adopts.

\section{Methodology}

\subsection{Search method}
We describe TND-NAS in detail in this section. According to the customs, under the supernetwork framework, we take $w$ to stand for the weight parameters, and $\alpha$ to represent the architecture parameters. As shown in Fig. \ref{fig:Illustration}, the search process is formulated as bilevel optimization problem w.r.t $\alpha$ and $\omega$. 
During the search, TND-NAS resorts to the continuous relaxation of the search space, the progressively shrinking of the candidate operations (presented as the ``search space approximation'' in \cite{chen2019progressive}), and first-order approximation of differentiable NAS. Importantly, the architecture parameters $\alpha$ are not directly trained by gradient descent with differentiable loss function (cross-entropy loss), but rather trained by reinforcement learning (REINFORCE \cite{williams1992simple}) in discrete space, which follows the non-differentiable route (marked blue in Fig. \ref{fig:Illustration}). Whereas we keep the optimization of $\omega$ in the differentiable route (marked red in Fig. \ref{fig:Illustration}). 

\subsubsection{Training of weights $w$}
Commonly, the supernetwork is constructed by the stack of mixed-edge operation $o$.
In terms of the weight parameters training, $w$ is optimized by the gradient descent:
\begin{equation} \label{eq:train_loss}
\begin{aligned}
{\mathop{w}}^{*}(\alpha) = \mathop{\arg\min}_{\mathop{w}}\mathcal{L}_{train}(\mathop{w},\alpha)
\end{aligned}
\end{equation}

\subsubsection{Sampling architectures by $\alpha$}
In terms of the architecture sampling, for each index $(i,j)$, $p^{(i,j)}$ is calculated by softmax of ${\alpha}^{(i,j)}$, the $g^{(i,j)}$ is sampled by multinominal distribution with probability vector $p^{(i,j)}$.
\begin{equation}
\begin{aligned}
p^{(i,j)} = softmax(\alpha^{(i,j)})
\end{aligned}
\end{equation}

\begin{equation}
\begin{aligned}
g^{(i,j)} \sim Multi(p^{(i,j)}, 1)
\end{aligned}
\end{equation}
the sub-network structured by $g$ inherits the supernetwork's weight parameters:
\begin{equation}
\begin{aligned}
N_{m} = \mathcal{A}(g_{m})
\end{aligned}
\end{equation}


\subsubsection{Optimized by REINFORCE}
We empirically resort to the REINFORCE with baseline \cite{williams1992simple}, in the process of gradient ascent, the gradient is estimated by sampling, that is:
\begin{equation} \label{eq:expect_reward}
\begin{aligned}
\nabla_{\alpha}{\mathcal{J}_{val}(\alpha)}&=\mathbb{E}_{N\sim\pi_{\alpha}(N)}[R_{val}({\mathop{w}}^{*},N)\nabla_{\alpha}\log(\pi_{\alpha}(N))]\\
&\approx\frac{1}{M}\sum_{m=1}^{M}(R_{val}({\mathop{w}}^{*},N_{m})-b)\nabla_{\alpha}\log(\pi_{\alpha}(N_{m}))
\end{aligned}
\end{equation}
where $M$ is the sample number of candidate architectures in one iteration. The baseline function is utilized to reduce the variance to reach the unbiased estimation of gradient, in which $b$ is the moving average of the previous architecture rewards.

\subsubsection{Reward calculation}
\begin{figure}[!ht]
\centering
\includegraphics[width=8.5cm]{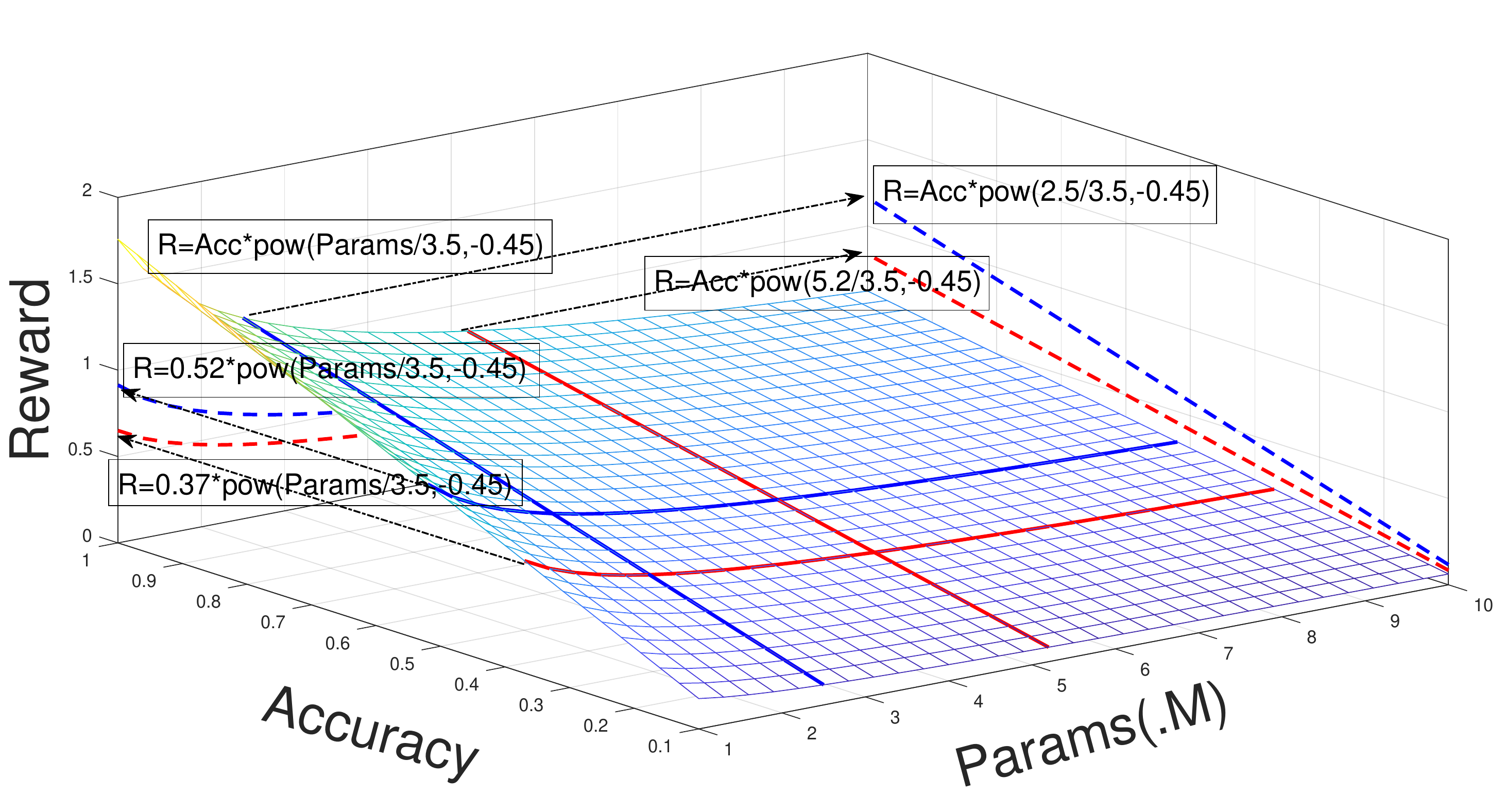}  
\caption{The surface of the reward function. 
}
\label{fig:reward_penalty_plot}
\end{figure}

Much more attention needs to be addressed to the multi-objective scenario, in which the reward function needs to be designed based on real-world requirements. For example, resource-constrained scenarios need the trade-off between performance and efficiency, e.g., memory consumption (mode size and number of accesses), or inference latency. Motivated by Mnasnet \cite{tan2019mnasnet}, taking the \textit{Accuracy} and \textit{Parameters} as the objectives, we make the reward be linearly w.r.t \textit{Accuracy}, but non-linearly w.r.t \textit{Parameters}, which is treated as a reward-penalty factor in the non-linearly scalarization function, presented in Eq. \eqref{eq:R-SO}. The reward function is shown in Fig. \ref{fig:reward_penalty_plot}.
\begin{equation}\label{eq:R-SO}
\begin{aligned}
&R=Acc \cdot (\frac{Params}{P})^{\beta} \\
\end{aligned}
\end{equation}

In this way, the non-differentiable metrics can naturally be involved in the search process by a scalarization function $f$ that transform the reward vector to a scalar one (single-policy MORL \cite{1998multi-criteria,mannor2004a,moffaert2013scalarized}).

We present the overall search method of TND-NAS in Algorithm \ref{algo:TND}.
\begin{algorithm}
\label{algo:TND}
\caption{TND-NAS}
\LinesNumbered
\For{$k = 0 \to stage\_num-1$}{
\text{Create a mixed operation } $\bar{o}_{k}^{(i, j)} \in \mathop{O_k^{(i,j)}}$\text{ parametrized by } $\alpha_{k}^{(i, j)}$ \text { for each edge }$(i, j)$;\\
Increase layer number $L_{k}$ by Eq. \eqref{eq:Layer};\\
Init the super-network by $\bar{o}_{k}^{(i, j)}$ and $L_{k}$;\\
\For{$epoch = 0 \to num\_epoch-1$}{
\For{$step = 0 \to train\_step$}{
\If {epoch \textgreater $e_k$}{\CommentSty{/*Pre-training for $e_k$ epochs*/}\\
\If {step\ \%\ rl\_interval == 0}{
\text{Update architecture } $\alpha$ \text { by} Eq. \eqref{eq:expect_reward}\\
}
}
\text{Update weights} $\omega$  \text{by} Eq. \eqref{eq:train_loss}\\
}
}
Shrinking each $\mathop{O}_k^{i,j}$ by Eq. \eqref{eq:O};\\
}
\text{Deriving the final architecture.}
\end{algorithm}

\section{Experiments}
\label{sec:Experiments}
\subsection{Datasets}
Our search and training processes are conducted on 2 popular image classification datasets, CIFAR10 and CIFAR100 \cite{krizhevsky2009learning}.

\subsection{Architecture Search}
\subsubsection{Experimental setting}
We conduct our experiments using PyTorch $1.4$ framework on 2 NVIDIA $1080Ti$ GPUs that each with $11GB$ memory. 

\subsubsection{Search space}
In terms of the predefined backbone of the search space, we follow the DARTS/P-DARTS, but as for the operations set, we make slight adjustments.\\ 
Normal cell:
\begin{tasks}(3)
\task none
\task skip\_connect
\task sep\_conv\_3x3
\task sep\_conv\_5x5
\task sep\_conv\_7x7
\task dil\_conv\_3x3
\task dil\_conv\_5x5
\task conv 1x1
\task conv 3x3
\task conv\_3x1\_1x3
\end{tasks}
Reduction cell:
\begin{tasks}(3)
\task none
\task skip\_connect
\task max\_pool\_3x3
\task avg\_pool\_3x3
\task max\_pool\_5x5
\task max\_pool\_7x7
\end{tasks}

\subsubsection{Search results}
\begin{figure}[!htb]
\centering
\subfigure[Normal cell.] {         
\includegraphics[width=6.5cm]{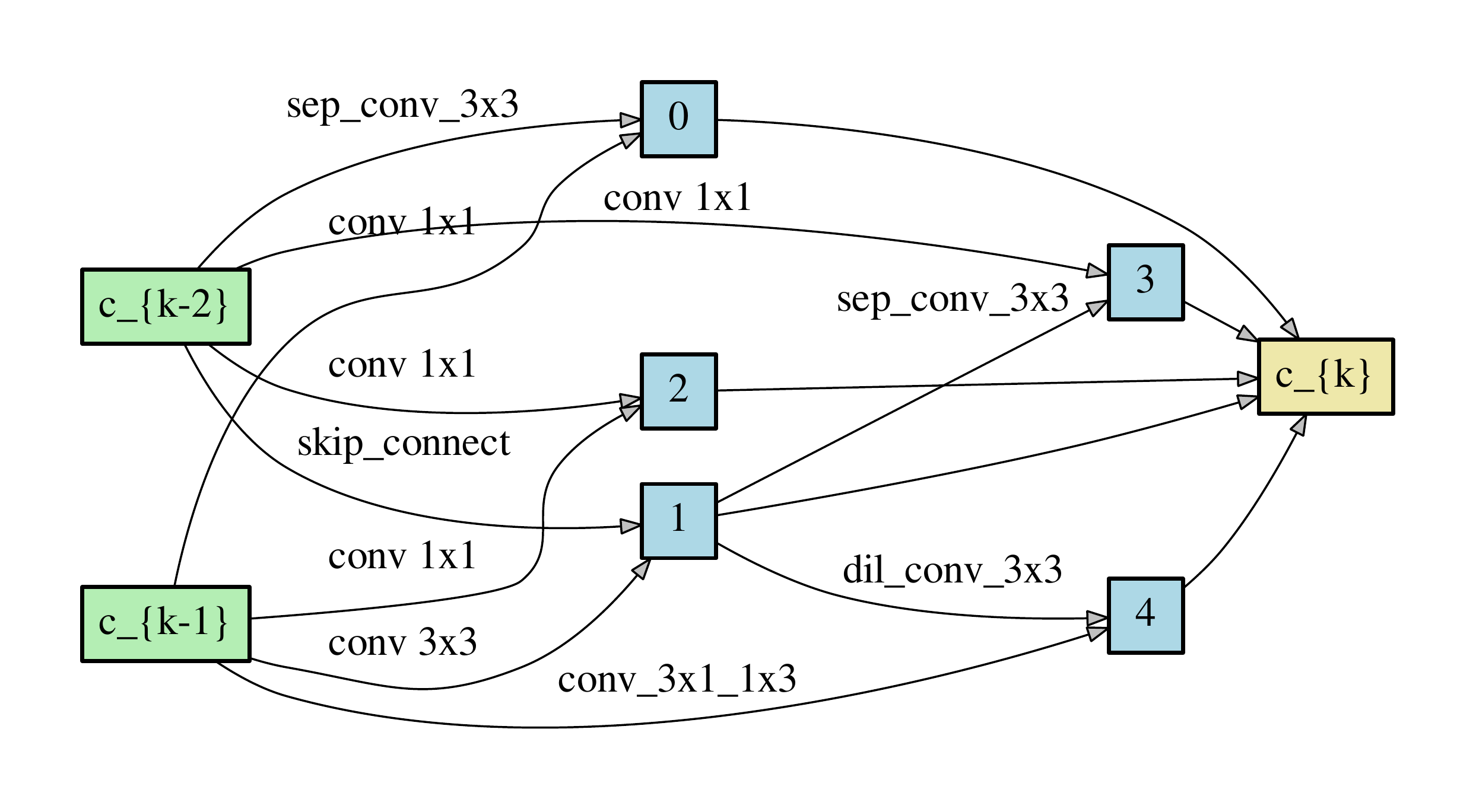}
\label{fig:normal_cell}
}
\subfigure[Reduction cell.] {         
\includegraphics[width=6.5cm]{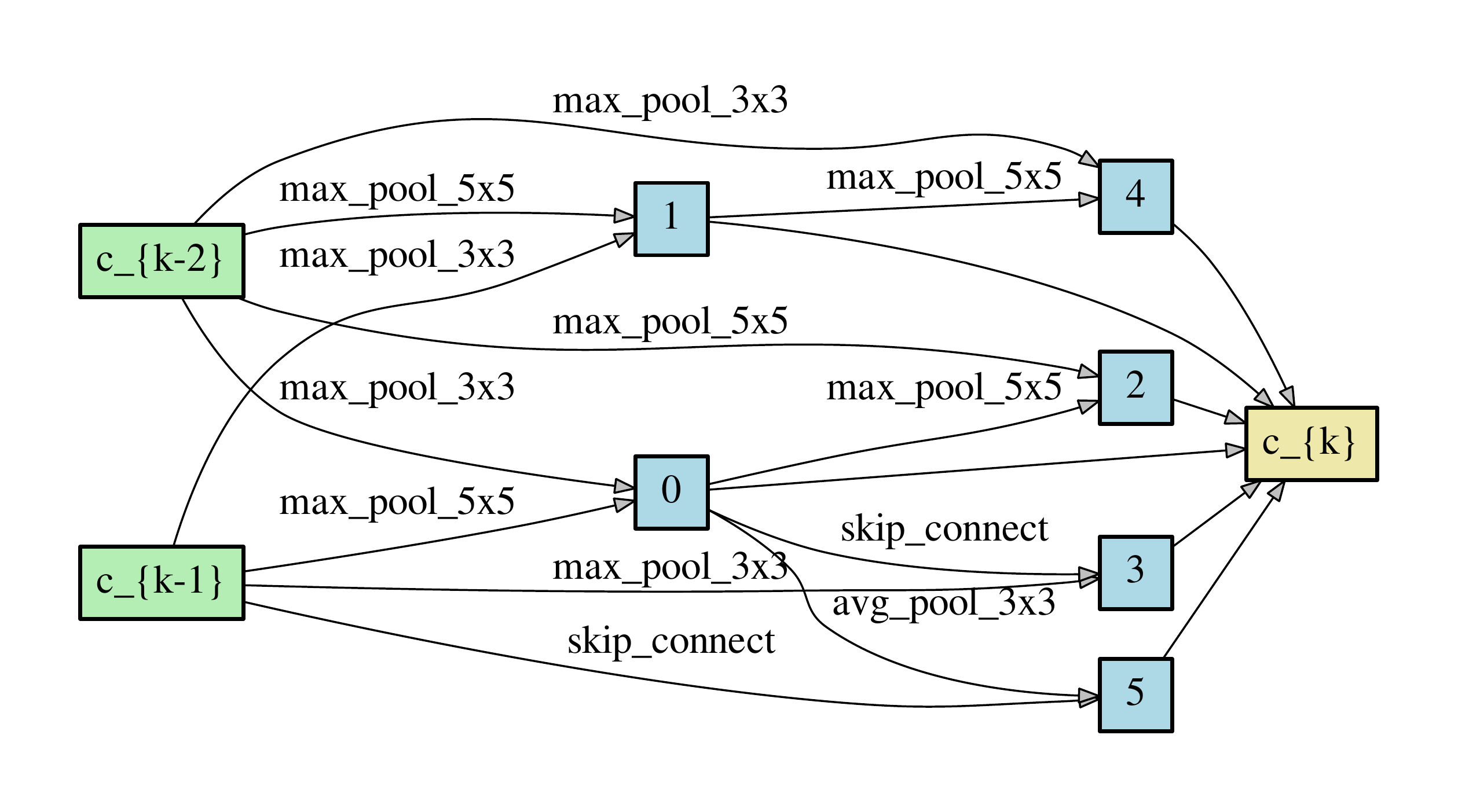}
\label{fig:reduction_cell}
}
\caption{The searched result (CIFAR10-S) of TND-NAS. 
}
\label{fig:searchresult}
\end{figure}

Our searched normal cell's visualization is shown in Fig. \ref{fig:normal_cell}, and reduction cell is shown in Fig. \ref{fig:reduction_cell}. Our search process costs merely 0.65 days on 2 NVIDIA 1080Ti GPU, each with only 11G memory. Compared with previous promising multi-objective NAS methods, our method achieves a substantial improvement in search resource cost (1.3 GPU-days). 

\begin{table}[]
\caption{Search cost comparison with state-of-the-art multi-objective NAS methods.}
\begin{tabular}{lll}
\hline
Methods & Search strategy & Search cost \\ \hline
MONAS \cite{hsu2018monas}& RL & / \\
MnasNet \cite{tan2019mnasnet}& RL & 4.5 days on 64 TPUv2 \\
LEMONADE \cite{elsken2019efficient}& EA & 80 GPU days \\
DPP-Net \cite{dong2018dpp-net}& SMBO & 8 GPU days on 1080Ti \\
PARETO-NASH \cite{elsken2018moas}& EA & 56 GPU days \\
lu2019nsga & EA & 8 GPU days on 1080Ti\\
TND-NAS & gradient+RL & \textbf{1.3 GPU days on 1080Ti} \\ \hline
\end{tabular}
\end{table}

\subsubsection{Results analysis}
From Fig. \ref{fig:searchresult}, the deep connection is preserved, and $conv\ 1\times1$ is frequently selected for its compact in \textit{Parameters} and the ability of feature fusion among channels of the feature map.

\subsection{Architecture Evaluation}
\subsubsection{Training details}
Our training details follow the experimental setting of evaluation in P-DARTS.

\subsubsection{Evaluation on CIFAR10 and CIFAR100}
\begin{table*}[!htb]
\caption{Comparison of the evaluation results.}
\label{table:performance}
\setlength{\tabcolsep}{5.5mm}
\begin{tabular}{lccccc}
\hline
\multirow{2}{*}{Architecture} & \multicolumn{2}{l}{Test Err. (\%)} & \multirow{2}{*}{\begin{tabular}[c]{@{}l@{}}Params\\ (M)\end{tabular}} & \multirow{2}{*}{\begin{tabular}[c]{@{}l@{}}Search Cost\\ (GPU-days)\end{tabular}} & \multirow{2}{*}{Search Method} \\ \cline{2-3}
 & C10 & C100 &  &  &  \\ \hline
DenseNet-BC \cite{huang2017densely} & 3.46 & 17.18 & 25.6 & - & manual \\
NASNet-A + cutout \cite{zoph2018learning} & 2.65 & - & 3.3 & 1800 & RL \\
AmoebaNet-A + cutout \cite{real2019regularized} & 3.34 & - & 3.2 & 3150 & evolution \\
AmoebaNet-B + cutout \cite{real2019regularized} & 2.55 & - & 2.8 & 3150 & evolution \\
Hireachical Evolution \cite{liu2017hierarchical} & 3.75 & - & 15.7 & 300 & evolution \\
PNAS \cite{liu2018progressive} & 3.41 & - & 3.2 & 225 & SMBO \\
ENAS + cutout \cite{pham2018efficient} & 2.89 & - & 4.6 & 0.5 & RL \\ \hline
DARTS (first order) \cite{liu2018darts} & 3 & 17.76 & 3.3 & 1.5 & gradient-based \\
DARTS (second order) + cutout \cite{liu2018darts} & 2.76 & 17.54 & 3.3 & 4 & gradient-based \\
SNAS + mild constraint + cutout \cite{xie2018snas}& 2.98 & - & 2.9 & 1.5 & gradient-based \\
SNAS + moderate constraint + cutout \cite{xie2018snas}& 2.85 & - & 2.8 & 1.5 & gradient-based \\
ProxylessNAS + cutout \cite{cai2018proxylessnas}& 2.08 & - & 5.7 & 4 & gradient-based \\ \hline
P-DARTS CIFAR10 + cutout \cite{chen2019progressive}& 2.5 & 16.55 & 3.4 & 0.3 & gradient-based \\
P-DARTS CIFAR100 + cutout \cite{chen2019progressive}& 2.62 & 15.92 & 3.6 & 0.3 & gradient-based \\
P-DARTS CIFAR10 (large) + cutout \cite{chen2019progressive}& 2.25 & 15.27 & 10.5 & 0.3 & gradient-based \\
P-DARTS CIFAR100 (large) + cutout \cite{chen2019progressive}& 2.43 & 14.64 & 11 & 0.3 & gradient-based \\ \hline
TND-NAS CIFAR10 (S) + cutout & 3.3 & - & 1.09 & 1.3$^\dagger$ & gradient + RL \\
TND-NAS CIFAR10 (M)+ cutout & 2.70 & - & 3.2 & 1.3$^\dagger$ & gradient + RL \\
TND-NAS CIFAR10 (L)+ cutout  & 2.54 & - & 9.57 & 1.3$^\dagger$ & gradient + RL \\
TND-NAS CIFAR100 (S)+ cutout & -  & 18.3 & 2.46 & 1.3$^\dagger$ & gradient + RL \\ 
TND-NAS CIFAR100 (M)+ cutout & -  & 16.73& 5.46 & 1.3$^\dagger$ & gradient + RL \\ 
TND-NAS CIFAR100 (L)+ cutout & -  & 15.20& 12.88 & 1.3$^\dagger$ & gradient + RL \\ \hline
\end{tabular}\\
\footnotesize{$^\dagger$ Performed on 2 Nvidia 1080Ti GPU each with 11G memory for 0.65 day}\\
\end{table*}

For comparison, some state-of-the-art approaches are listed in Table. \ref{table:performance}.
Our experiment reaches a series of scalable models on CIFAR10.
Our search experiment on CIFAR100 also achieves the promising results.

\section{Ablation and Diagnostic Experiments}
\subsection{Effect of search algorithm}
\subsubsection{Reinforcement of performance}
\begin{figure*}[!ht]
    \centering
        \includegraphics[width=16.5cm]{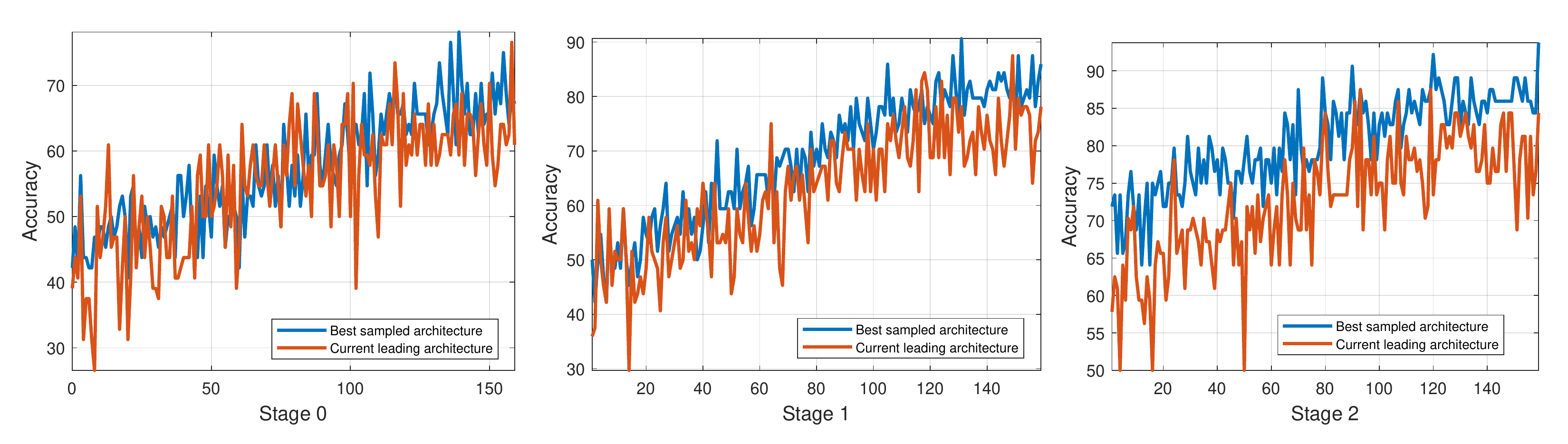}
    \caption{Reinforcement effect of searching for the best-performance architecture. The $max(Acc(\mathcal{A}(g)))$ and $Acc(\mathcal{A}(argmax(\alpha)))$ increase synergistically.}
        \label{fig:00_multinominal_and_max}
\end{figure*}
First, we verify the effectiveness of the search algorithm on the differentiable metric. 
As can be seen from Fig. \ref{fig:00_multinominal_and_max}, the two precision values are continuously improved, which reflects the effectiveness of the algorithm in effectively searching out the architectures with excellent precision performance.

\subsubsection{Reinforcement of \textit{Parameters}}
\begin{figure*}[!ht]
    \centering
        \includegraphics[width=16.5cm]{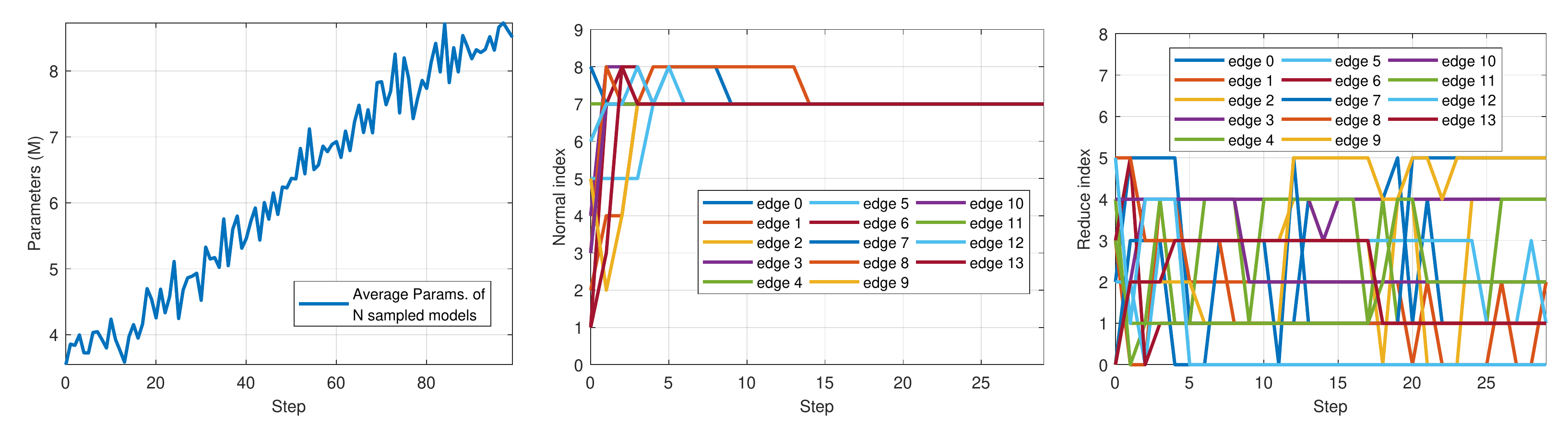}
    \caption{Reinforcement effect to search for the maximum \textit{Parameters} architecture. The average \textit{Parameters} of the sampled models continuously increase.
    }
        \label{fig:max_and_min_para}
\end{figure*}

\begin{figure*}[!ht]
    \centering
        \includegraphics[width=16.5cm]{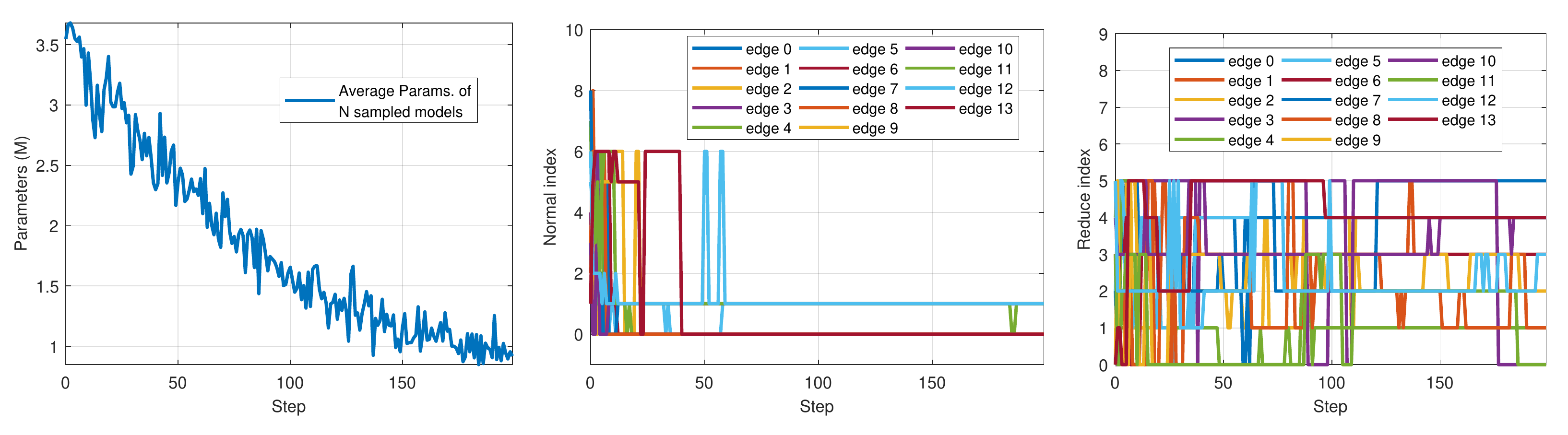}
    \caption{Reinforcement effect of searching for the minimum \textit{Parameters} architecture. The average \textit{Parameters} of the sampled models continuously decrease, the Normal cell architecture that determine the \textit{Parameters} also converges to $skip\_connection$ (index 1) and $none$ operation (index 0)}
        \label{fig:max_and_min_para}
\end{figure*}


\begin{figure*}[!ht]
    \centering
        \includegraphics[width=17.5cm]{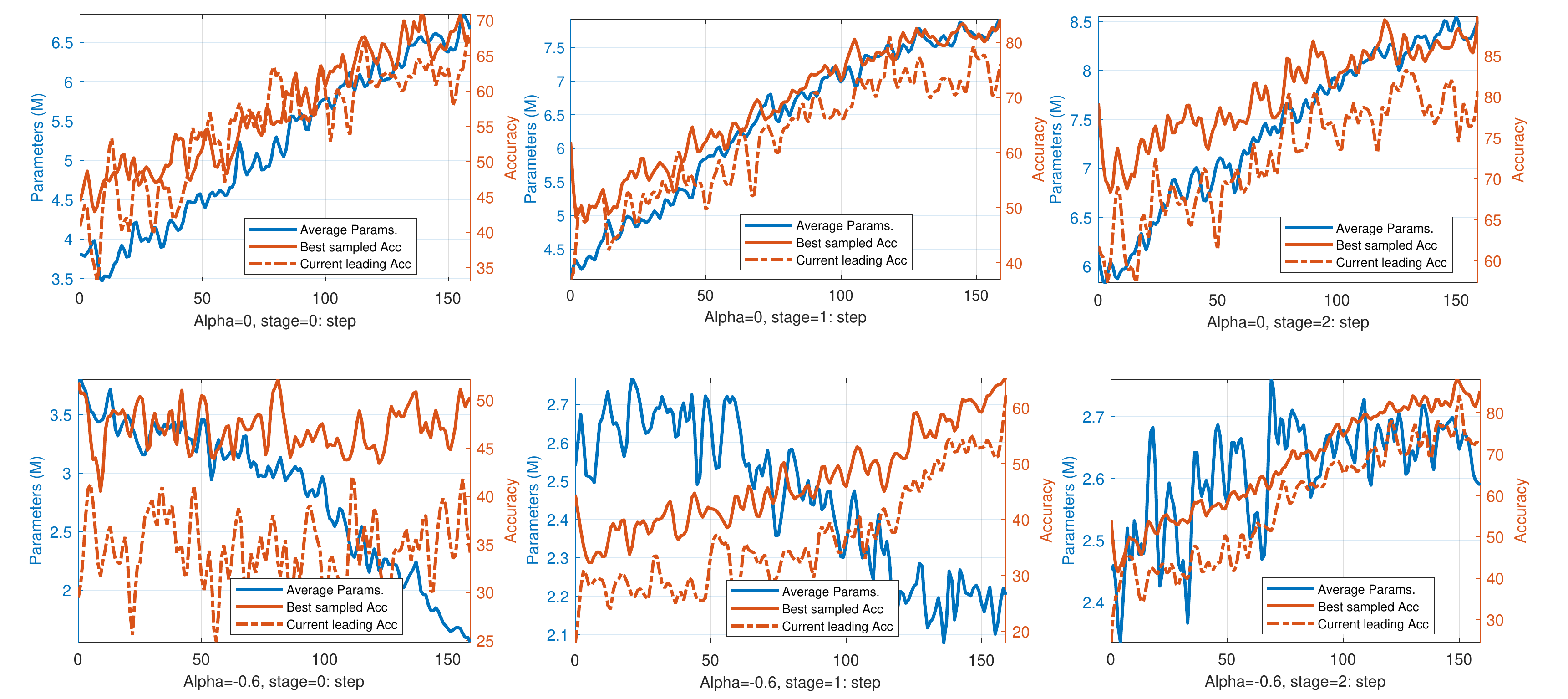}
    \caption{Effect of the model compression without performance sacrifice, following the reward presented in Eq.(9). Top: penalty coefficients of 0, the increase in \textit{Accuracy} at the cost of the increase in \textit{Parameters}. Bottom: penalty coefficients of -0.25), the \textit{Parameters} are maintained, but the \textit{Accuracy} is still enhanced.}
        \label{fig:compress_compare}
\end{figure*}
\subsection{Effect of the model compression without performance sacrifice}
We further focus on the effect of the search that incorporates the metrics of accuracy and computational cost. Our experiment performs the comparison between two different penalty coefficients (0 and -0.25) by tracking the \textit{Parameters} and \textit{Accuracy}. It can be seen that, in the experiment with penalty coefficients of 0, the \textit{Parameters} and \textit{Accuracy} values increase synergistically, which can be interpreted as that the increase in \textit{Accuracy} at the cost of the increase in \textit{Parameters}. However, after employing the reward-penalty mechanism (with penalty coefficients of -0.25), while the \textit{Parameters} are maintained and the accuracy is still enhanced.

\section{Conclusion}
\label{sec:conclusion}
This work incorporates the differentiable NAS framework with the capability to handle non-differentiable metrics, and aims to reach the trade-off among non-differentiable and differentiable metrics.
Meanwhile, our method reconciles the merits of multi-objective NAS and differentiable NAS and is feasible to the applied in the real-world NAS scenarios, e.g. resource-constrained, and platform-specialized.
Taking the \textit{Parameters} as an example, after the multi-objective search, we achieved a series of scalable and promising models. 
Favorably, it only costs 1.3 GPU-days on \textit{NVIDIA 1080Ti}, which is 1/6 of that in NSGA-Net.
Since our work aims at proposing the novel NAS methods, but not the intentional design of the transferability, we do not carry out further discussion and experiments on this.
Further due to the limitation of computational resources and our motivation, we do not conduct the search experiment and full training on ImageNet, and the representative experiments do not include some other non-differentiable metrics, e.g. \textit{inference latency}.
While the gaps and non-differential metrics issues have been addressed, it inevitably leads to a greater computational cost.
Targeting the specialized NAS scenarios, the search hyperparameters setting (reference, penalty coefficient) of each metrics is necessary, so the self-adaptive tuning of these hyperparameters is a topic worthy of research. Further, it is necessary to improve the efficiency of the search framework.


\bibliographystyle{IEEEtran}
\bibliography{ref}

\end{document}